\documentclass[11pt,reqno,twoside]{amsart}

\usepackage{amsfonts}
\usepackage{amsmath,amssymb,amsthm,amsthm}
\usepackage{mathtools}
\usepackage{dsfont}
\usepackage{etoolbox}
\usepackage{color}
\usepackage{graphicx,color}
\usepackage{dsfont,subfigure}
\usepackage{fixmath}
\usepackage{tcolorbox}
\usepackage{algorithm}
\usepackage{algpseudocode}
\usepackage{graphicx}
\usepackage{enumerate}
\usepackage{tabularx}

\usepackage[dvips,bottom=1.4in,right=1in,top=1in, left=1in]{geometry}

\makeatletter
\def\eqnarray{\stepcounter{equation}\let\@currentlabel=\theequation
\global\@eqnswtrue
\tabskip\@centering\let\\=\@eqncr
$$\halign to \displaywidth\bgroup\hfil\global\@eqcnt\z@
  $\displaystyle\tabskip\z@{##}$&\global\@eqcnt\@ne
  \hfil$\displaystyle{{}##{}}$\hfil
  &\global\@eqcnt\tw@ $\displaystyle{##}$\hfil
  \tabskip\@centering&\llap{##}\tabskip\z@\cr}

\def\endeqnarray{\@@eqncr\egroup
      \global\advance\c@equation\m@ne$$\global\@ignoretrue}

\title[Novel DNNs for Stiff ODEs with Applications to Chemically Reacting Flows]{Novel DNNs for Stiff ODEs with Applications to \\ Chemically Reacting Flows}

\author{Thomas S. Brown$^{1,3}$, Harbir Antil$^{3}$, Rainald L\"ohner$^1$, 
        Fumiya Togashi$^2$, \\
        Deepanshu Verma$^{3}$}
\address{$^1$Center for Computational Fluid Dynamics,
             College of Science, George Mason University, 
             Fairfax, VA 22030-4444, USA.}
\email{tbrown62@gmu.edu,rlohner@gmu.edu}               

\address{$^2$Applied Simulations, Inc.,
             1211 Pine Hill Road, McLean, VA 22101, USA}
\email{fumiya.togashi@gmail.com}               
               
\address{$^3$Center for Mathematics and Artificial Intelligence (CMAI),
             College of Science, 
             George Mason University,
             Fairfax, VA 22030-4444, USA.}
\email{hantil@gmu.edu, dverma2@gmu.edu}

\thanks{This work was supported by the Defense Threat
Reduction Agency (DTRA) under contract HDTRA1-15-1-0068.
Jacqueline Bell served as the technical monitor.}

\begin{document}

\begin{abstract}
Chemically reacting flows are common in engineering, such as hypersonic 
flow, combustion, explosions, manufacturing processes and environmental 
assessments. For combustion, the number of reactions can be significant 
(over 100) and due to the very large CPU requirements of 
chemical reactions (over 99\%) a large number of flow and combustion 
problems are presently beyond the capabilities of even the largest 
supercomputers.  \\
Motivated by this, novel Deep Neural Networks (DNNs) are introduced to 
approximate stiff ODEs. Two approaches are compared, i.e., either learn 
the solution or the derivative of the solution to these ODEs. 
These DNNs are applied to multiple species and reactions common in chemically 
reacting flows. Experimental results show that it is helpful to account 
for the physical properties of species while designing DNNs. 
The proposed approach is shown to generalize well.
\end{abstract}

\maketitle

%%%%%%%%%%%%%%%%%%%%%%%%%%
\section{Introduction}\label{s:intro}
%%%%%%%%%%%%%%%%%%%%%%%%%%

Chemically reacting flows are common in many fields of engineering, such as 
hypersonic flow, combustion, explosions, manufacturing processes, and 
environmental assessments \cite{A124,stuck2010adjoint,J85,J93}. For 
hydrocarbon combustion and explosions the numbers of species and reactions 
can reach into hundreds and thousands respectively. Even with so-called 
reduced models \cite{VaVaTu2004, Keck1990, MaPo1992, LaGo1994, LuLa2005, SuChGoJu2010}, which try to keep the main species and reactions while neglecting 
those that %either release minor amounts of energy or 
are not important,
%for the reactions to happen, 
typically over 100 reactions need to be updated. An expedient (and widely 
used) way to compute flows with chemical reactions is to separate the 
advection and diffusion of species from the actual reactions. In this way, 
the vastly different timescales of the reactants can be treated in a 
separate, stiff ODE solver. Such chemical reaction solvers take the given 
species ${u}^{n-1}$ at the $n-1^{\rm th}$ time step and desired 
timestep $\delta t$ and update the species to ${u}^{n}$. In terms of 
a `black box solver' this implies either: 
	\begin{equation}\label{eq:chem}
		{u}^{n} = Chem_1 (\delta t, {u}^{n-1}) \, ,
	\end{equation}
or 	
	\begin{equation}\label{eq:chem2}
		\frac{u^{n}-u^{n-1}}{\delta t} = Chem_2(u^{n-1}) ,
	\end{equation}
where $Chem_1$ stands for the ODE integrator of chemical reactions and 
$Chem_2$ is the right-hand side of the system. This is the formulation to 
solve the system of ODEs numerically using the standard Euler method. 
Compared to a typical `cold' flow case, the presence of these chemical 
reactions implies an increase of computing requirements that can exceed 
factors of 1:100, i.e. 2 orders of magnitude. This makes many of these 
flow simulations so expensive that entire classes of problems have been 
sidelined, waiting for faster computers to be developed in years to come.
The goal here is to replace the `black box' solvers (equivalently the 
functions $\emph{Chem}_1$ and $\emph{Chem}_2$) given in \eqref{eq:chem} 
and \eqref{eq:chem2} by novel, robust Deep Neural Networks (DNNs) without 
sacrificing accuracy. 

The list of references on using DNNs in computational fluid dynamics (CFD) 
is growing fast, see for example, \cite{LyMiRa2020, GrFa2020}. However, 
the results on using DNNs in chemical kinetics are scarce. A popular 
approach to solve PDEs and ODEs is through the use of so-called 
Physics-Informed Neural Networks (PINNs) \cite{RaPeKa2019, ChZh2021}. The 
goal of PINNs is to minimize the PDE/ODE residual by using a neural network 
as a PDE/ODE solution Ansatz. The inputs to the network are space-time 
variables $(x,t)$ and all the derivatives are computed using automatic 
differentiation. See  \cite{JiQiShPaDe2020} for an example of a PINN for 
stiff ODE systems where the only input is time. 

The approach presented here fundamentally differs from the aforementioned 
approaches. Instead of \emph{Physics-Informed-Neural-Networks}, the goal 
is to pursue  \emph{Learn-from-Physics/Chemistry}. For instance in 
\eqref{eq:chem} and \eqref{eq:chem2}, DNNs will be used to 
learn $Chem_1$ and $Chem_2$ from a given dataset coming from physics/chemistry 
simulations. Such an approach to learn $Chem_1$ has also been considered 
recently in \cite{PePi2017, ShJoKeMo2020, OwPa2021} where the authors 
employ an architecture that is motivated by standard feed forward networks. 
The authors of \cite{ZhEtAl2020} consider a similar problem, but use an 
autoencoder, which is a type of DNN used to reduce the dimension of the system. 
% In order to achieve the intended goal of this work, novel DNNs are introduced, instead of using the existing ones.
Notice that the proposed approach will allow for the chemical reactions 
described by \eqref{eq:chem} and \eqref{eq:chem2} to start at any point 
in time without knowing a reference time. The latter is crucial for the 
application under consideration. 

%In this paper, we introduce 
The DNNs used in this paper have been motivated by the Residual Neural 
Network (ResNet) architecture. ResNets have been  
introduced in  \cite{HeZhReSu2016,RuHa2020, HaKhLoVe2020} in the context of 
data/image classification, see also \cite{antil2021novel} for
parameterized PDEs and \cite{GhEtAl2020} where the (related) so-called 
Neural ODE Nets \cite{ChRuBeDu2018} have been used to solve
stiff ODEs. The ResNet architecture is known to overcome the vanishing 
gradient problem, which has been further analyzed using fractional 
order derivatives in \cite{HaKhLoVe2020}. The key feature of a ResNet is 
that in the continuum limit, it becomes an optimization problem constrained 
by an ODE. Such a continuous representation further enables the analysis of 
the stability of the network using nonlinear ODE theory. In addition, 
standard approaches from optimization, such as the Lagrangian approach can 
be used to derive the sensitivities with respect to the unknown weights 
and biases. 

The main novelties in the DNNs presented in this work are the following: 
\begin{itemize}
\item These networks allow for learning both the solution 
(see \eqref{eq:chem}) and difference quotients (see \eqref{eq:chem2}).
A similar approach to learn the solution in the context of parameterized 
PDEs has been recently considered
in \cite{antil2021novel}. 
\item Motivated by chemically reacting flows, the goal is to create 
networks that can learn multiple reactions propagating multiple species. 
To accomplish this task, parallel ResNets are constructed where the data 
corresponding to multiple quantities is used as input for each network but 
the output is only a single species. %We emphasize that 
Similar approaches for chemical kinetics can be found in 
\cite{PePi2017, ShJoKeMo2020}, where the authors use standard feed forward 
networks. 
\item The proposed DNNs are applied to non-trivial chemically reacting flows.
Experimental results show that it is helpful to know the underlying 
properties of the species while designing the networks.
\end{itemize}	

The remainder of the paper is organized as follows: In 
Section \ref{sec:ResNet}, the DNNs used to approximate systems of 
type \eqref{eq:chem} and \eqref{eq:chem2} are introduced, and training 
strategies are discussed.
Section \ref{sec:Imp} provides some information on the implementation of 
the method. Several experimental  results are discussed in 
Section \ref{sec:Res}. Additional information on the data being used in the 
experiments is provided in Appendix \ref{sec:app}. 

\section{Deep Residual Neural Networks} \label{sec:ResNet}% to Learn ODEs}

\subsection{Problem Formulation}

Consider an input-to-output map
	\[
		u \mapsto S(u), 
	\]
where $S$ could be $Chem_1$ or $Chem_2$ in \eqref{eq:chem} 
and \eqref{eq:chem2}, respectively. 
The goal is to learn an approximation $\widehat{S}$ of $S$ using Deep 
Neural Networks (DNNs). In particular, the
proposed DNNs are motivated by Deep Residual Neural Networks (ResNets). 
See \cite{RuHa2020,HaKhLoVe2020} for examples of ResNets for classification 
problems and \cite{antil2021novel} for an application to parameterized PDEs. 

\begin{figure}[htb]
	\centering
	\includegraphics[width=0.3\textwidth]{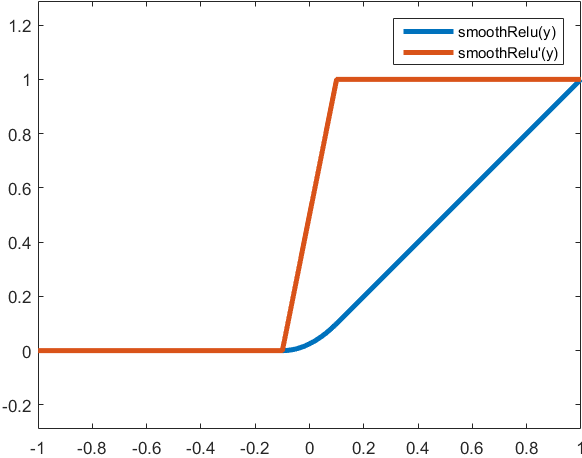} \quad 
	\includegraphics[width=0.55\textwidth]{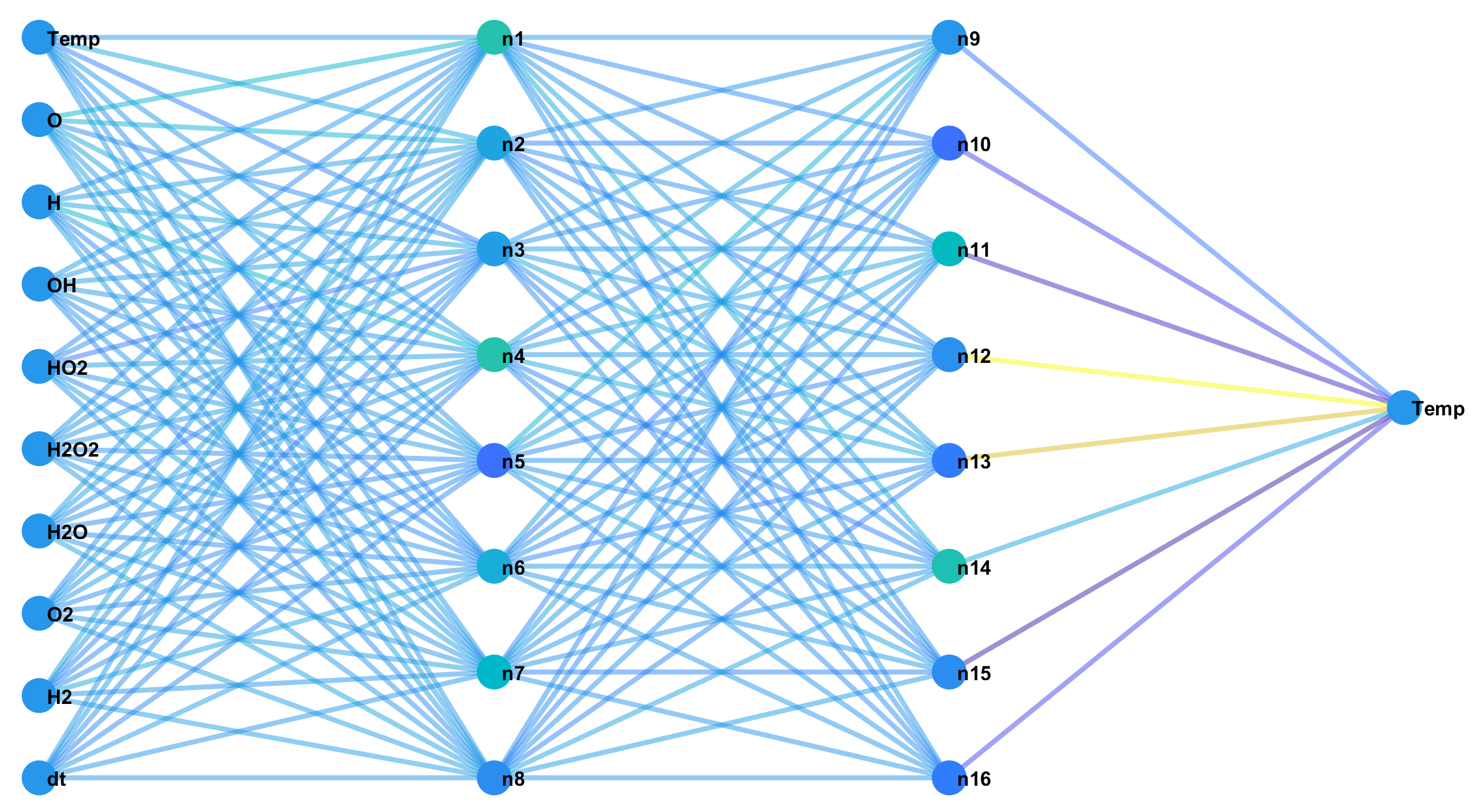}
    \includegraphics[width=0.036\textwidth]{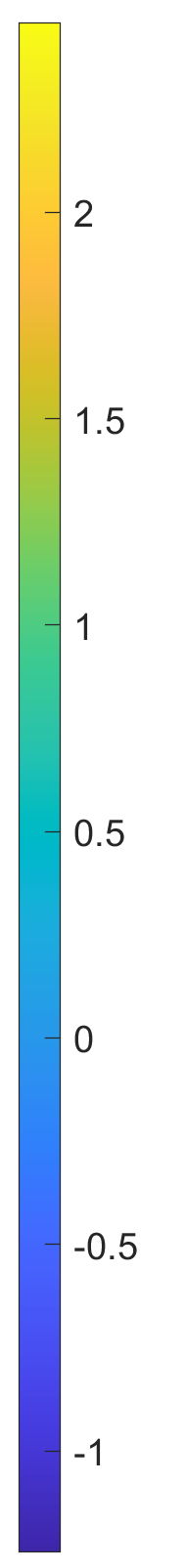}
	\caption{Left: Smooth ReLU and its derivative for $\varepsilon =0.1$. 
                 Right: Example of a typical Deep ResNet used in the experiments.}
	\label{f:relu}
\end{figure}

Given a training dataset $\{(u^n,S(u^n))\}_{n=0}^{N-1}$, the DNN 
approximation $\widehat{S}$ of $S$ is given by the output of the DNN (ResNet) 
%	\begin{equation} \label{eq:netRep}
%		\widehat{S}(u) = (y_L \circ y_{L-1} \circ \cdots \circ y_1)(u)
%	\end{equation}
%where $\{y_\ell\}_{\ell=1}^L$ is generated using the DNN (ResNet) 
	\begin{equation}\label{eq:resnet}
		\begin{cases}
			y_1 = \tau \sigma(K_0 y_0 + b_0) ,\\
			y_\ell = y_{\ell-1} + \tau \sigma( K_{\ell-1} y_{\ell-1}+ b_{\ell-1} ), \quad & 1 < \ell \le L-1 , \\
			y_L    = K_{L-1} y_{L-1},	
		\end{cases}
	\end{equation}
i.e., $\widehat{S}(u) = y_L$. Here 
%$y_0 = u^0$ and 
$\tau > 0$ is a given parameter called the skip connection parameter. 
The nonlinear function $\sigma$ denotes an activation function. In this 
study, the activation function is taken to be a smooth quadratic 
approximation of the ReLU function, i.e., 
	\[
		\sigma(x) = \begin{cases} 
\max\{0,x\}                                 & |x|>\varepsilon,\\
\frac1{4\varepsilon} x^2 + \frac12 x + \frac{\varepsilon}{4} 
                                            & |x|\leq \varepsilon.
		            \end{cases}		
	\]
Figure~\ref{f:relu} (left) shows that $\sigma(x)$ is a good approximation 
of ReLU. Here, and in the experiments below, $\varepsilon$ is taken to be 0.1. 
The number of layers (depth of the network) is denoted by $L$. 
Layer 0 is referred to as the input layer, layers 1 through $L-1$ as 
hidden layers and layer $L$ as the output layer.

The quantities $\{K_\ell\}_{\ell=0}^{L-1}, \{b_\ell\}_{\ell=0}^{L-2}$ 
denote the weights and biases, i.e. the parameters that need to be 
determined. In this setting $K_\ell$ is a matrix and $b_\ell$ is a vector, 
together they introduce an affine transformation. 
If $y_\ell \in \mathbb R^{n_\ell}$ for $\ell = 0, \dots, L$, 
then  $K_\ell \in \mathbb R^{n_{\ell+1} \times n_\ell}$ and 
$b_\ell \in \mathbb R^{n_{\ell + 1}}$. The dimension $n_\ell$ is also referred 
to as the width of the $\ell$-th layer.  Notice that for $ 0 < \ell < L$ 
in \eqref{eq:resnet}, it is assumed that $n_\ell = n_{\ell +1}$, but this 
can be easily generalized, see \cite{RuHa2020}. Nevertheless, in the current 
setup the dimension of the input $y_0$ and output $y_L$ can be different.  

%The notation $n_\ell$ for $1 \leq \ell \leq L-1$ is used as the width of each layer. 
An example of a typical DNN is shown in Figure \ref{f:relu} (right) with 
depth 3, width 8, input dimension 10, and output dimension 1.  The values 
of the weights are given by the color of the lines connecting the neurons, 
the values of the biases are given by the color of the neurons, and the 
color of the input layer has been set to zero.   

The question remains of how to compute the weights 
$\{K_\ell\}_{\ell=0}^{L-1}$ and biases $\{b_\ell\}_{\ell=0}^{L-2}$ 
in order to obtain a good approximation $\widehat{S}$ of $S$. 
Following \cite{RuHa2020,HaKhLoVe2020}, these weights are computed by solving 
	\begin{equation}\label{eq:opt}
		\min_{\{K_\ell\}_{\ell=1}^{L-1}, \{b_\ell\}_{\ell=0}^{L-2}} \left\{ J(K_\ell,b_\ell) := \frac{1}{2N} \sum_{n=0}^{N-1} \| y_L^n -  S(u^n) \|^2 \right\} \quad \mbox{subject to constraints} \quad	
		\eqref{eq:resnet} , 
	\end{equation}
where $y_L^n = \widehat S(u^n)$, that is, $y_0$ in \eqref{eq:resnet} is taken to be the training data $\{u^n\}_{n=0}^{N-1}$.
The problem \eqref{eq:opt} is a constrained optimization problem. To solve 
this problem, the gradient of $J$ with respect to $K_\ell$ and $b_\ell$ 
needs to be evaluated. This requires introducing the so-called adjoint 
equation (also known as back-propagation in machine learning literature). 
See \cite{RuHa2020,HaKhLoVe2020,antil2021novel} for complete details. Having obtained
the gradient, an approximate Hessian is computed via the BFGS routine.
Both the gradient and the approximate Hessian are used to solve the 
minimization problem.

\subsection{Proposed Approach and Applications} \label{sec:Approach}

As stated earlier, the goal is to replace $Chem_1$ and $Chem_2$ 
in \eqref{eq:chem} and \eqref{eq:chem2}, respectively by DNNs. 
%We begin by rewriting these equations. 
The continuous ODE is given by 
%Our goal is to use a Dense Neural Net with ResNet structure as an ODE solver.  We use two different approaches to achieve this goal. First, consider an ODE
	\begin{equation}\label{eq:ode}
		\frac{du}{dt}(t) = S(u(t)) , \quad u(0) = u_0.
	\end{equation}
The simplest time discretization of \eqref{eq:ode} is given by 
	\begin{equation}\label{eq:ode1}
		\frac{u^n - u^{n-1}}{\delta t} = S(u^{n-1}) , \quad u^0 = u_0 ,
	\end{equation}
for $n=1,\dots, N$.  
%DNNs are used to approximate the right-hand-side of the ODE with $\widehat{S}$, and 
At first, the right-hand-side of \eqref{eq:ode1} is approximated. 
This is referred to as the \emph{first approach}.
In this case, the training data (input and output) is 
$\left\{\left(u^{n-1}, \frac{u^n - u^{n-1}}{\delta t} \right) \right\}_{n=1}^{N}$.  
%We will refer to this as approach number one. 
In the \emph{second approach}, the map to be approximated is 
$u^{n+1} = S(u^n) $, i.e., the setting of \eqref{eq:chem}. Here $S$ 
marches $u$ forward in time with time increments given by $\delta t$. 
The training data for this approach is given by 
$\{(u^{n-1}, u^{n})\}_{n=1}^{N}$.

\paragraph{ \bf Scaling.}  
The training data in the examples below has been computed using 
CHEMKIN \cite{KeEtAl2000}.  Before training the networks, the data is 
scaled in the following manner:  For a data set $\{x_j\}_{j=0}^N$ 
corresponding to a single quantity (e.g. temperature) the mean 
$\mu_x$ and standard deviation $\sigma_x$ of the set are computed.  
Each entry $x_i$ is scaled as 
\begin{equation} \label{eq:scale}
z_i := \frac{x_i - \mu_x}{\sigma_x} ~~.
\end{equation}
Using this new centered and normalized set, the training data is 
obtained by defining
\begin{equation} \label{eq:scale2}
\widehat{x}_i := \frac{z_i- \min_j z_j}{\max_j z_j - \min_j z_j}, 
\end{equation}
so that the resulting data set $\{\widehat{x}_j\}_{j=1}^N$ lies in the 
interval $[0,1]$. Given that chemical reactions follow an exponential
Arrhenius-type law, a logarithmic scaling was also tried. This implies
performing the above scaling on the dataset $\{\log x_j\}$ instead of 
$\{x_j\}$.

\paragraph{ \bf Architecture.} For the DNNs implemented below, the 
input dimension will always be given by 
\begin{alignat*}{15}
n_0 =& &&1 &&+  && M  && +  &&1  &&= M+2.\\
 & (\text{temp} && \text{erature})\qquad  && \qquad (\text{\# of } && \text{species}) \qquad && \qquad (\text{time inc}&&\text{rement, $\delta t$})
\end{alignat*}
The output dimension is  $n_L = M+1$, which corresponds to temperature 
plus the number of species.

Rather than using a single DNN to approximate the entire solution map $S$, 
in many cases a parallel ResNet architecture is implemented in which a 
separate ResNet is created to correspond to each desired output. With 
this structure the output dimension for each ResNet is 1, but there 
are $M+1$ ResNets implemented in parallel. The inputs to all of the 
parallel ResNets are the same, and so the parallel architecture can also 
be thought of as a single large ResNet (with $n_L = M+1$) that is not 
fully connected.

\paragraph{ \bf Loss Function and Training.} In the case of a 
parallel ResNet architecture, each parallel network is associated to a 
separate loss function.  The same form of the loss function is used for 
each network.  Letting $\theta^{(i)}$ represent the concatenation of 
all weight and bias parameters associated to the $i$-th parallel network, 
the loss function takes the form 
\begin{equation}\label{eq:loss}
	\frac{1}{2N} \sum_{n=0}^{N-1} \|y_L^{(i)} - S(u^n)^{(i)}\|_2^2 + \frac{\lambda}{2} \left(\|\theta^{(i)}\|_1 + \|\theta^{(i)}\|_2^2\right), \qquad i = 1, \dots, M+1. 
\end{equation}
In other words, the process of training each network is the process 
of finding the parameters $\theta^{(i)}$ which minimize the mean squared 
error between the network output and the training data, while also using 
both $\ell^1$ and $\ell^2$ regularizations to penalize the size of the 
parameters.  As indicated in section \ref{sec:ResNet}, a gradient based 
method (BFGS in particular) is used to solve the constrained optimization 
problem.

\paragraph{ \bf Validation.} 
DNNs are trained with validation data in order 
to overcome the overfitting issue.  The validation data is a subset of the 
training data that is not used to compute the weights and biases. Instead, 
a separate loss function is formed that computes the mean squared error 
between the ResNet output and the validation data.  This is a way to test 
how the ResNet performs on unseen data during the training process itself.  
If the validation error increases, the training continues for a certain 
number of additional iterations, called the \emph{patience}.  During this 
time, if the validation error decreases to a new minimum, the patience 
resets to zero and training continues.
%\HA{to its initial value}. 
If the validation error does not attain a new minimum and the full number 
of patience iterations is reached, then the training process is terminated.

\paragraph{\bf Testing.} 
After training and validation comes the testing phase in which the DNN 
approximations are used to carry out time marching.
%we test the results by using the networks to solve a system of ODE by marching in time from an initial condition.  
Using the \emph{first approach} (see above) this involves implementing 
an Euler time-stepping commonly used to numerically solve ODEs, 
with the usual right-hand-side of the ODE replaced by the DNN output,
%except that where one would normally see the right-hand-side of the ODE system, the output of the DNN is used.  
that is, 
\begin{equation} \label{eq:EulerStep}
\widehat{u}^n = \widehat{u}^{n-1} + \delta t \widehat{S}(\widehat{u}^{n-1}) \qquad   n = 1,\dots, N,
\end{equation}
where $\widehat{u}^0 = u^0$ is known data.  For the \emph{second approach}, 
results are tested by using the following update step 
$\widehat{u}^n = \widehat{S}(\widehat{u}^{n-1})$ for $n = 1, \dots, N$, 
where again $\widehat{u}^0 = u^0$ is known.

\paragraph{\bf Application.} 
The above methods are applied to a system of ODEs that model 
hydrogen-oxygen reaction. In particular, the reduced hydrogen-air 
reaction model with 8 species and 18 reactions \cite{PeHa1999} is used. 
%Notice that one might expect 9 species in this case, however, we neglect nitrogen. 
%\todo{how do we justify ignoring nitrogen?}  
This model is simpler than the hydrocarbon reaction model mentioned in 
section~\ref{s:intro}. However, it can still capture several essential key 
features. More complicated DNNs which can handle over 100 reactions 
will be part of future work.

\section{Implementation} \label{sec:Imp}

To start, the \emph{second approach} was implemented in 
Keras \cite{ChEtAl2015} using stochastic gradient descent. 
Libraries such as Keras have proven to be highly successful for 
classification problems, but their success in physics based modeling is 
still limited. Initially, it was observed that the results produced by 
using Keras were unable to capture the `pulse' in chemical reactions. 
After gaining access to some of the Keras code, and seeing some improvement, 
the results were still highly inconsistent. For example, when running the 
same experiment multiple times, with the same 
hyperparameters, different solutions were obtained. Such an inconsistency may 
not be critical for classification problems 
(for example \cite{KrSuHi2017,LeBoBeHa1998}), but it is critical for 
ODE/PDE problems. Implementation through Keras was abandoned as further 
customization of the code proved to be a daunting task. 

The DNN implementation used in this paper was carried out in MATLAB 
and a BFGS routine with an Armijo line-search was used to solve the 
optimization problem. Unlike many neural network software packages, 
this code uses the framework and closely resembles code that is used to 
solve optimal control problems. 

\section{Results} \label{sec:Res}

In this section a variety of results are presented that represent 
different approaches and architectures. The results will be grouped 
together by the type of data that was used to train the networks.  Further 
details on the training data are provided in Appendix \ref{sec:app}. 
The loss function is as given in \eqref{eq:loss} with $\lambda = 1e-7$, 
the skip connection parameter (in \eqref{eq:resnet}) $\tau=2/(L-1)$,
where $L$ is the number of hidden layers, and a constant time increment of
$\delta t = 1e-7$.  Unless otherwise specified, all training data was 
scaled by the method outlined in \eqref{eq:scale} and \eqref{eq:scale2}.  
All of the results presented below were achieved by using the MATLAB 
implementation of the ResNets described above. The blue curves in the 
plots below represent `true solution' (data generated using CHEMKIN), 
while the dashed red curves represent DNN output.

\subsection{Training with a single data set.}

\begin{figure}[htb]
	\centering
	\includegraphics[width=0.45 \textwidth]{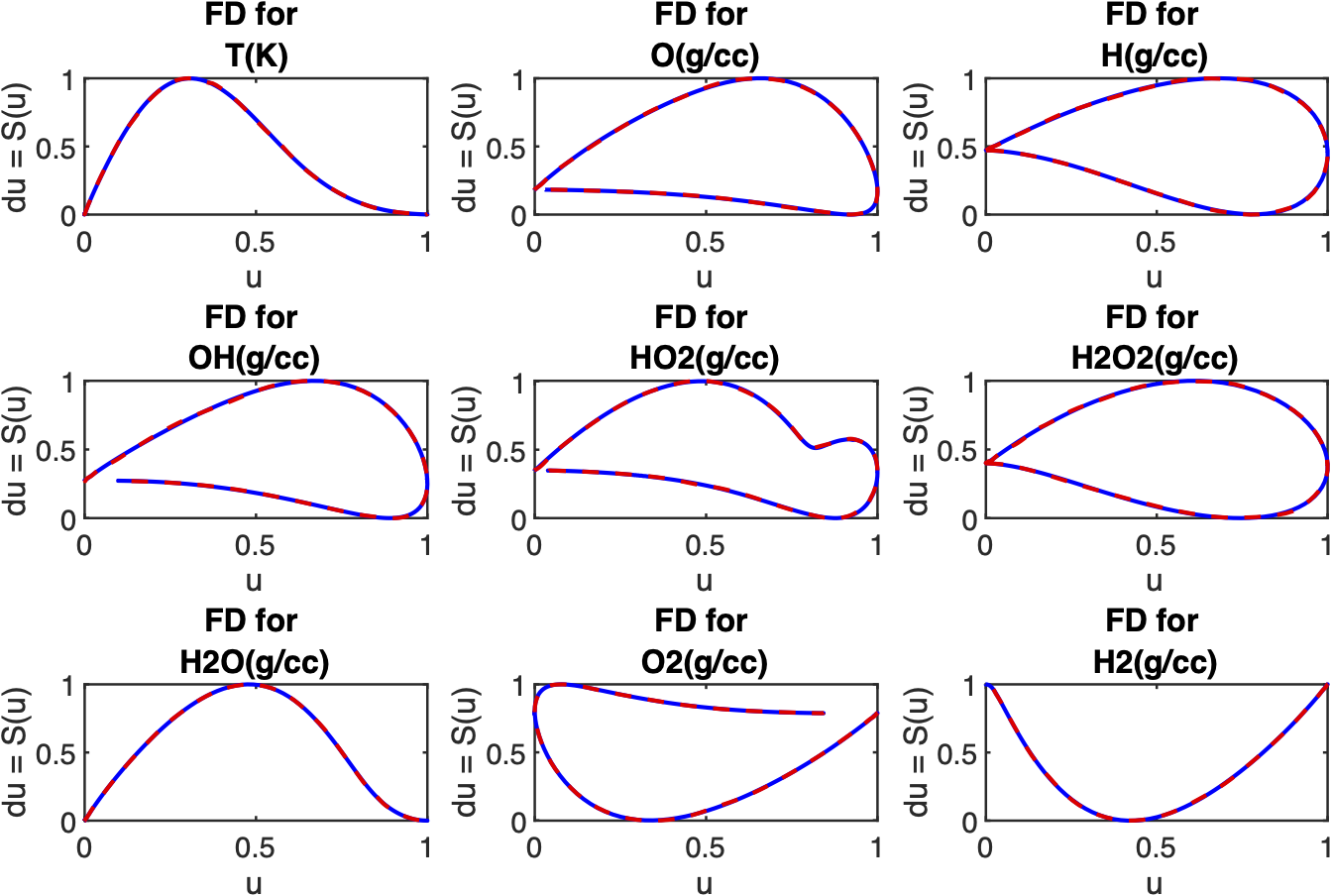} \quad   
 	\includegraphics[width=0.45\textwidth]{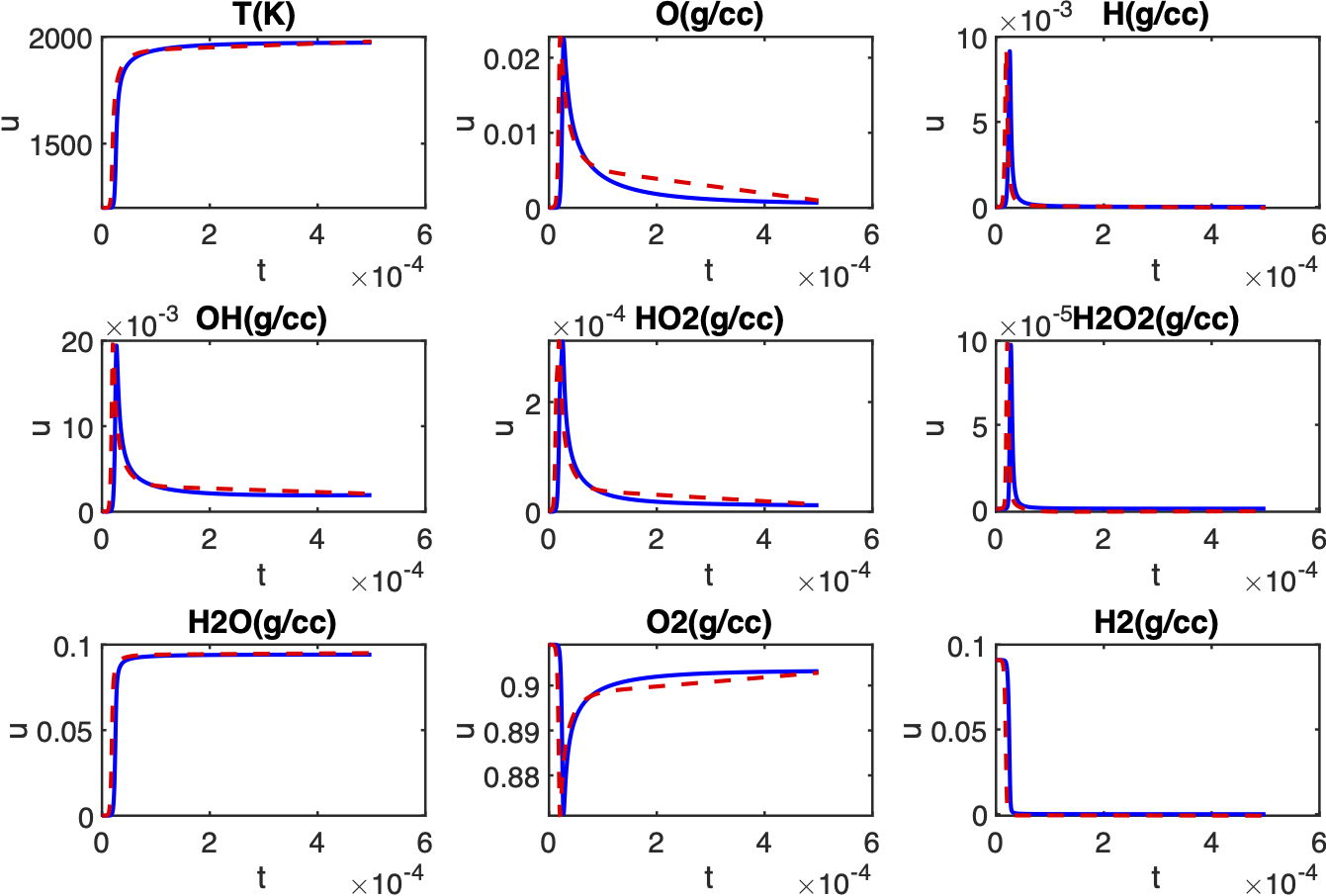}
	
	\caption{A comparison between the actual $du/dt$ and the learned difference quotient (left 9 plots) and the results of marching in time as in \eqref{eq:EulerStep} (right 9 plots) using a Deep ResNet of depth 9 and width 15.}
	\label{f:chem1}
\end{figure}

Using the \emph{first approach}, a single ResNet was trained on data 
corresponding to temperature, 8 species, and $\delta t$.  In 
Figure \ref{f:chem1}, the results from a ResNet with depth 9 and width 15 
are shown. For this experiment
%, the skip parameter is taken to be 0.25 and 
only $\ell^2$ regularization has been used in the loss function.  After 
being trained, the training data was used again to test the network, and 
these results can be seen in the left plots of Figure~\ref{f:chem1}.  
The initial condition from the training data was marched forward in time 
as described in \eqref{eq:EulerStep}, and the results are shown on the 
right of Figure \ref{f:chem1} compared with the actual $u$.

\subsection{Training data sets with varying equivalence ratio and 
fixed initial temperature.}

\begin{figure}[htb]
	\centering
	\includegraphics[width=0.48 \textwidth]{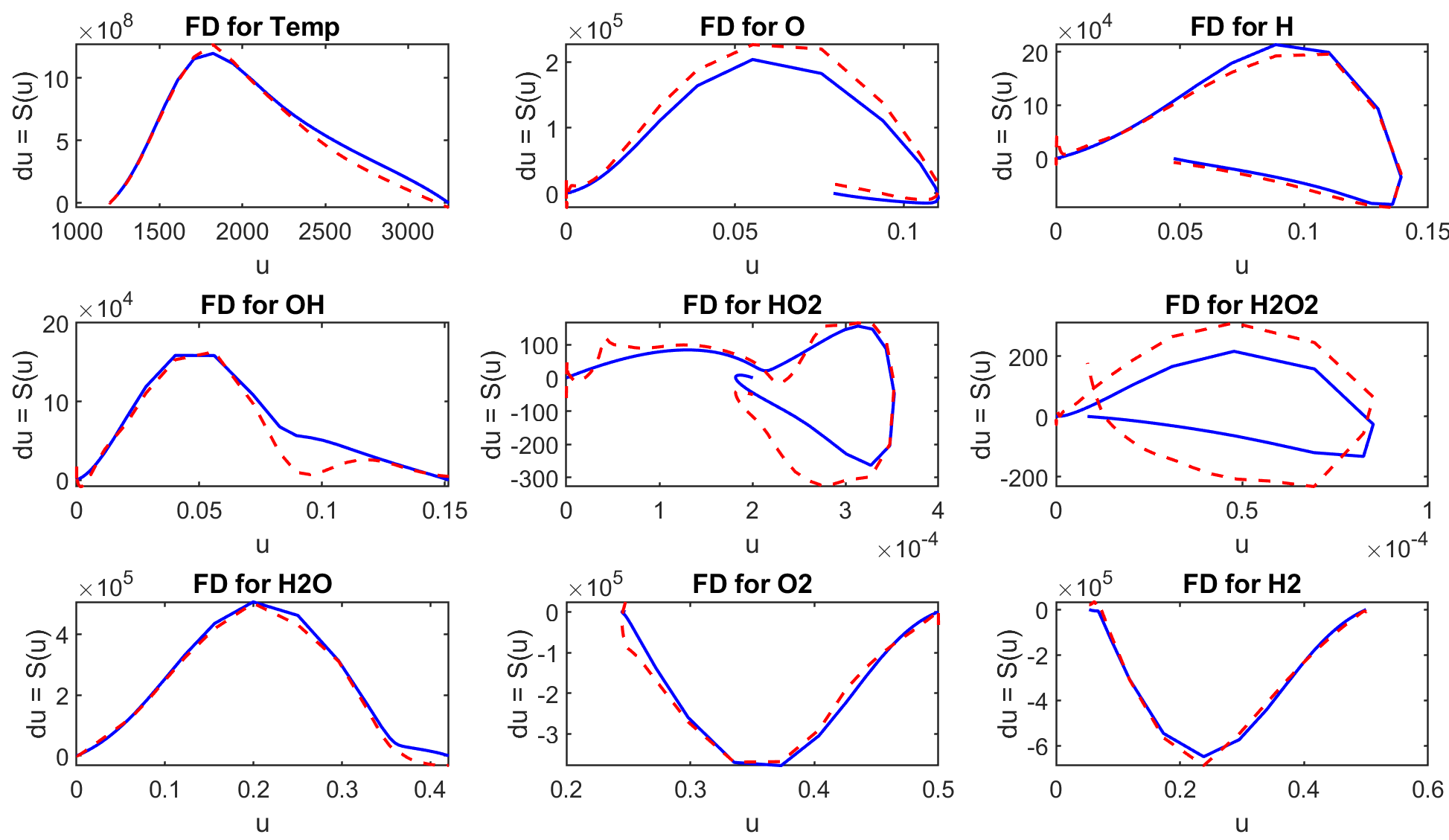} \quad   
 	\includegraphics[width=0.48\textwidth]{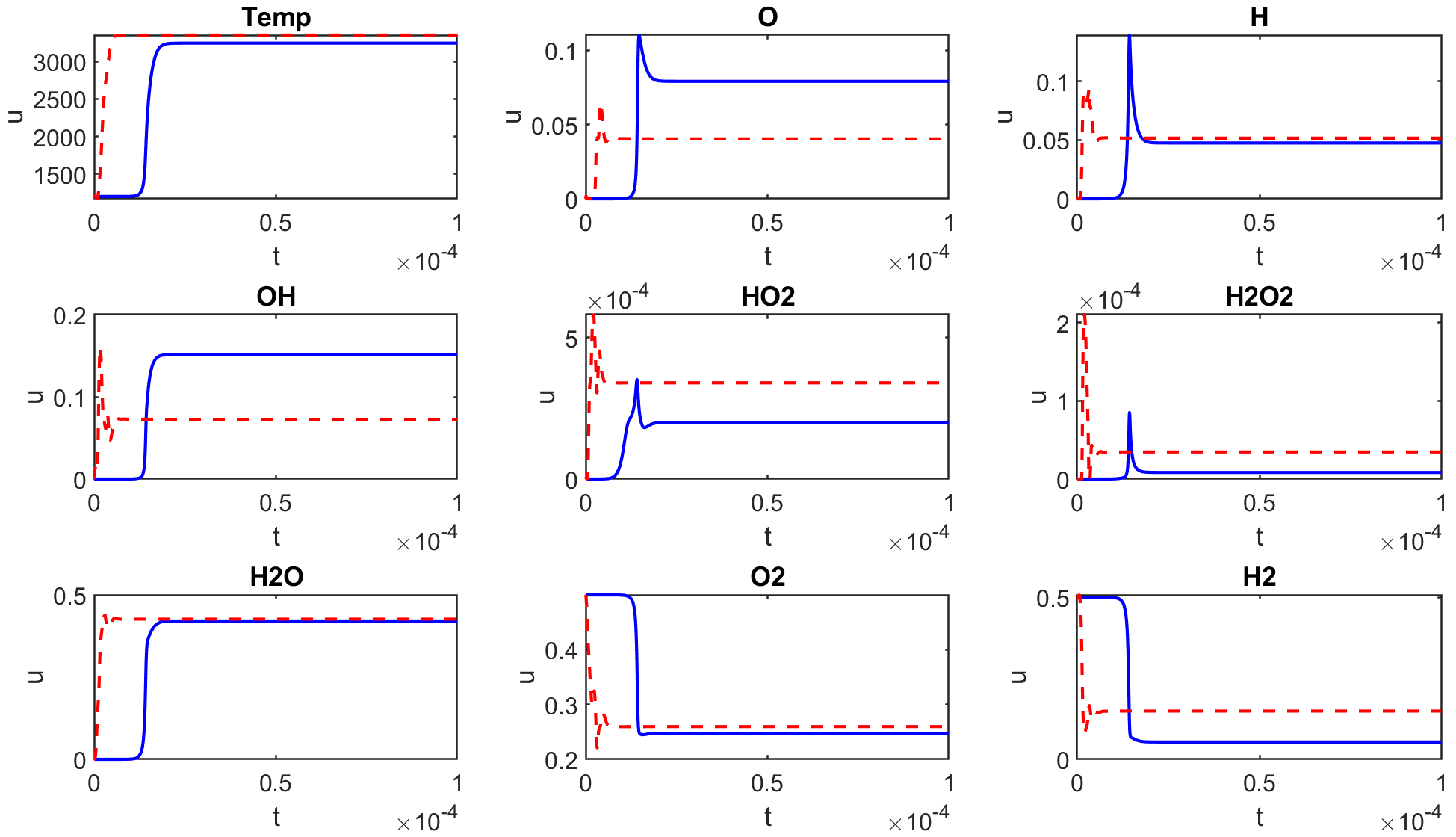}
	
	\caption{A comparison between the actual $du/dt$  and the learned difference quotient (left 9 plots), and the results of marching in time as in \eqref{eq:EulerStep} (right 9 plots) using parallel Deep ResNets of different depths and width 10, trained with validation data with patience of 250 iterations.}
	\label{fig:earlyValDat}
\end{figure}

Experimental results showed that while a single ResNet was able to handle 
data corresponding to a single initial condition, it struggled with data 
containing multiple initial conditions. To improve results, the parallel 
ResNet structure described in Section \ref{sec:Approach} was created.   
The parallel DNNs were trained on a data set containing multiple initial 
conditions with the same initial temperature but varying equivalence ratio 
(see Appendix \ref{sec:app}).  
Figure \ref{fig:earlyParSub} shows the results for ResNets trained on 
this data without the use of validation data.  Each ResNet has depth 4 
and width 10. For this result, the testing data was a subset of the 
training data, and so these plots do not represent a generalization of 
the networks to unseen data.
Figure \ref{fig:earlyValDat} shows the result of an experiment trained 
with a configuration of data subsets from the same data set as 
Figure \ref{fig:earlyParSub}, (see Appendix \ref{sec:app} for specific 
details).   Additionally, validation data was implemented during training 
with a patience of 250 iterations.  The ResNets used for this experiment 
have varying depths (3 to 5) but a fixed width of 10. The plots in 
Figure \ref{fig:earlyValDat} are the result of testing the DNNs on data 
that was not used in training. For these results mass conservation is also 
implemented in the sense that the sum of the densities of the 8 species 
is adjusted at the end of each update step \eqref{eq:EulerStep} to be 
equal to that of the previous step.  This is the only experiment presented 
in this work in which mass conservation is enforced.  

\begin{figure}[htb]
	\centering
	\includegraphics[width=0.48 \textwidth]{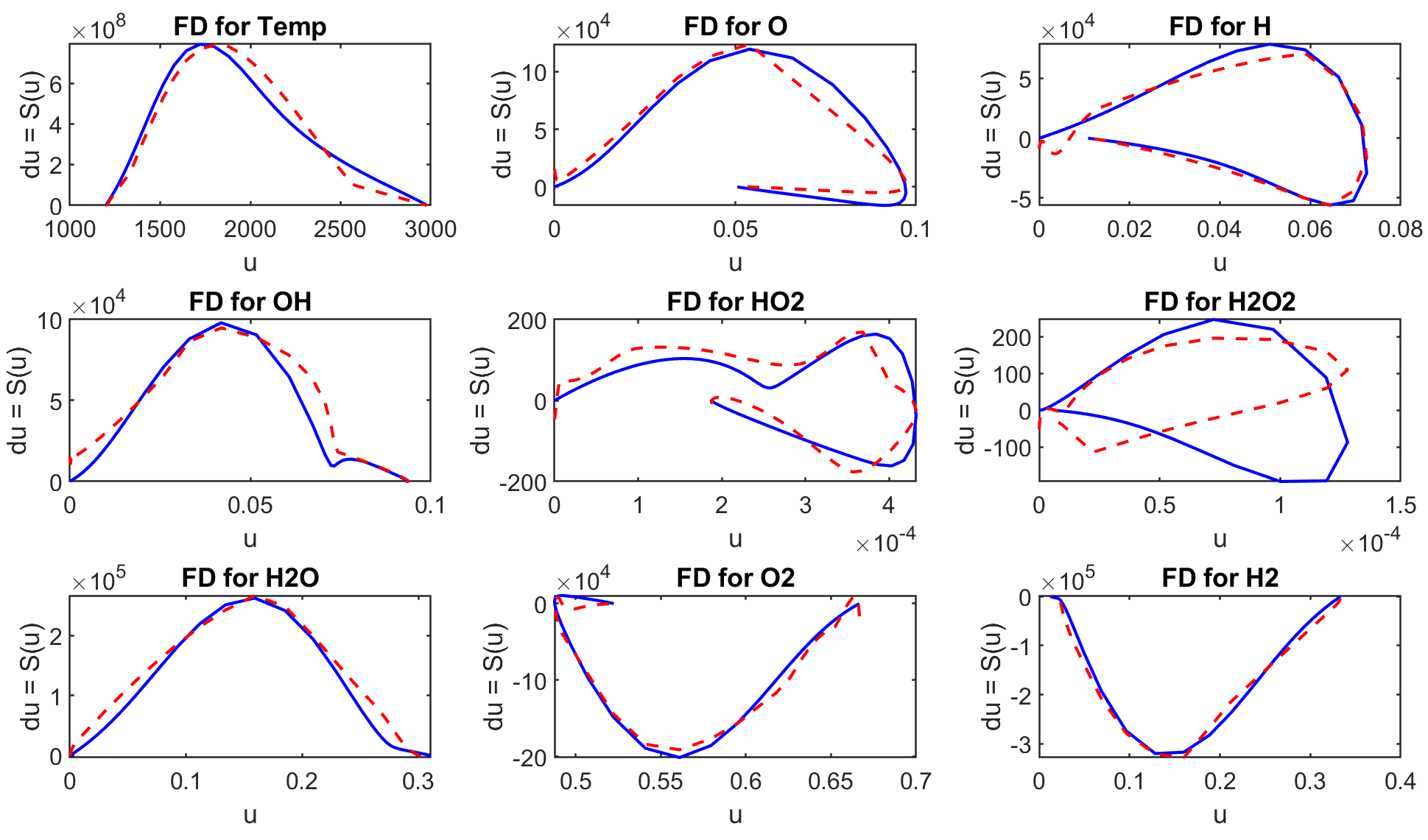} \quad   
 	\includegraphics[width=0.48\textwidth]{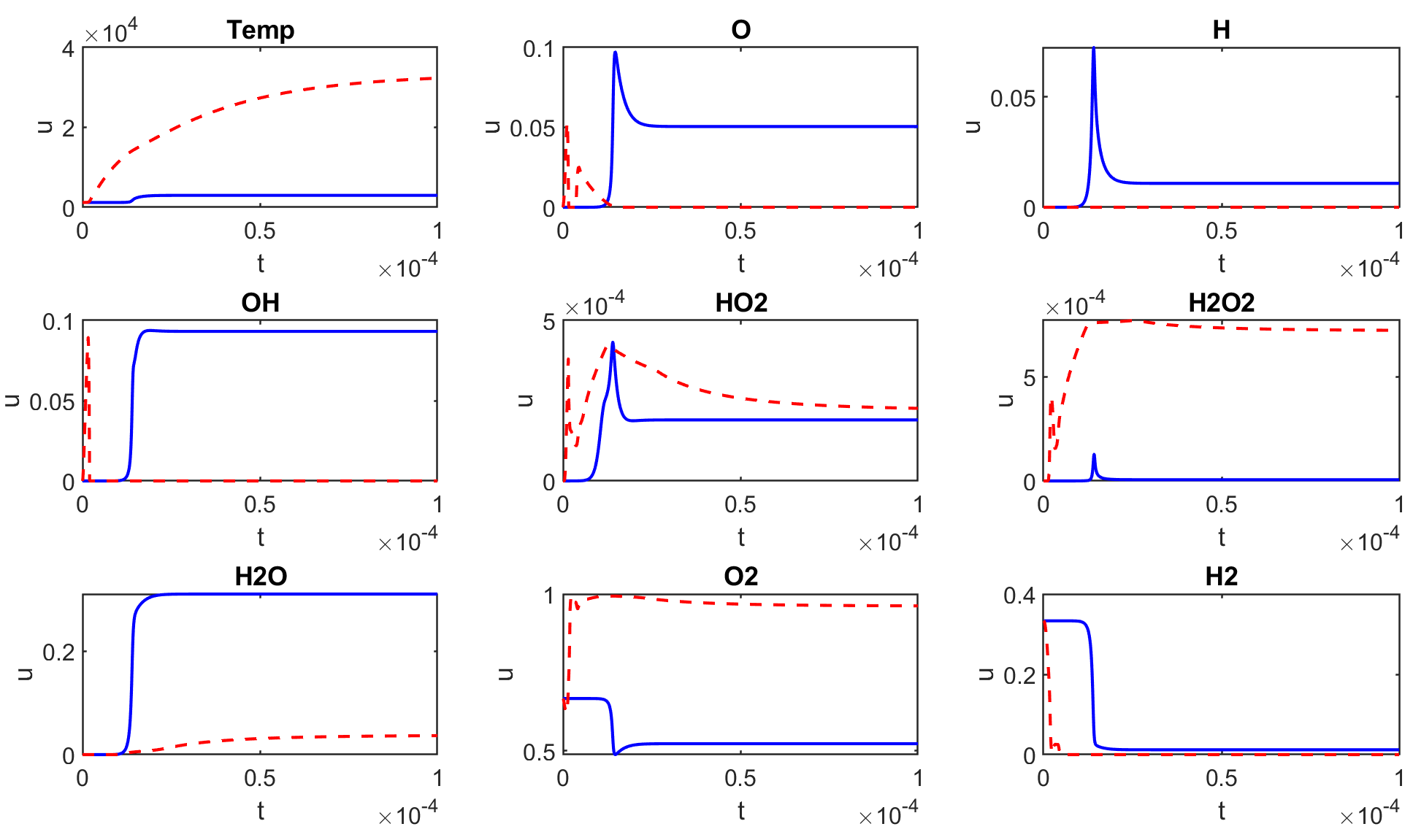}
	
	\caption{A comparison between the actual $du/dt$ and the learned difference quotient (left 9 plots), and  the results of marching in time as in \eqref{eq:EulerStep} (right 9 plots) using parallel Deep ResNets of depth 4 and width 10.}
	\label{fig:earlyParSub}
\end{figure}

\begin{figure}[htb]
	\centering
	\includegraphics[width=0.48 \textwidth]{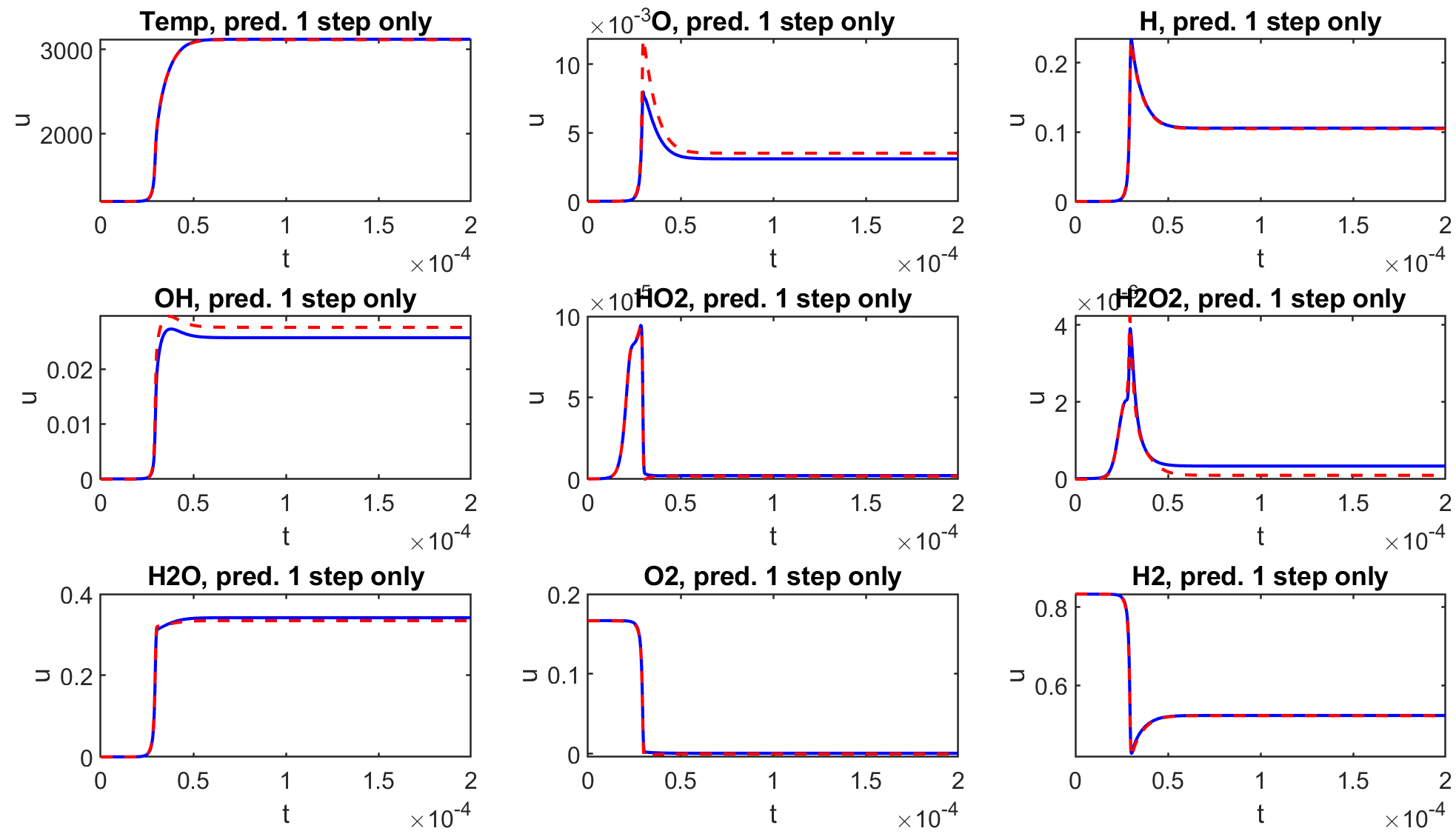} \quad   
 	\includegraphics[width=0.48\textwidth]{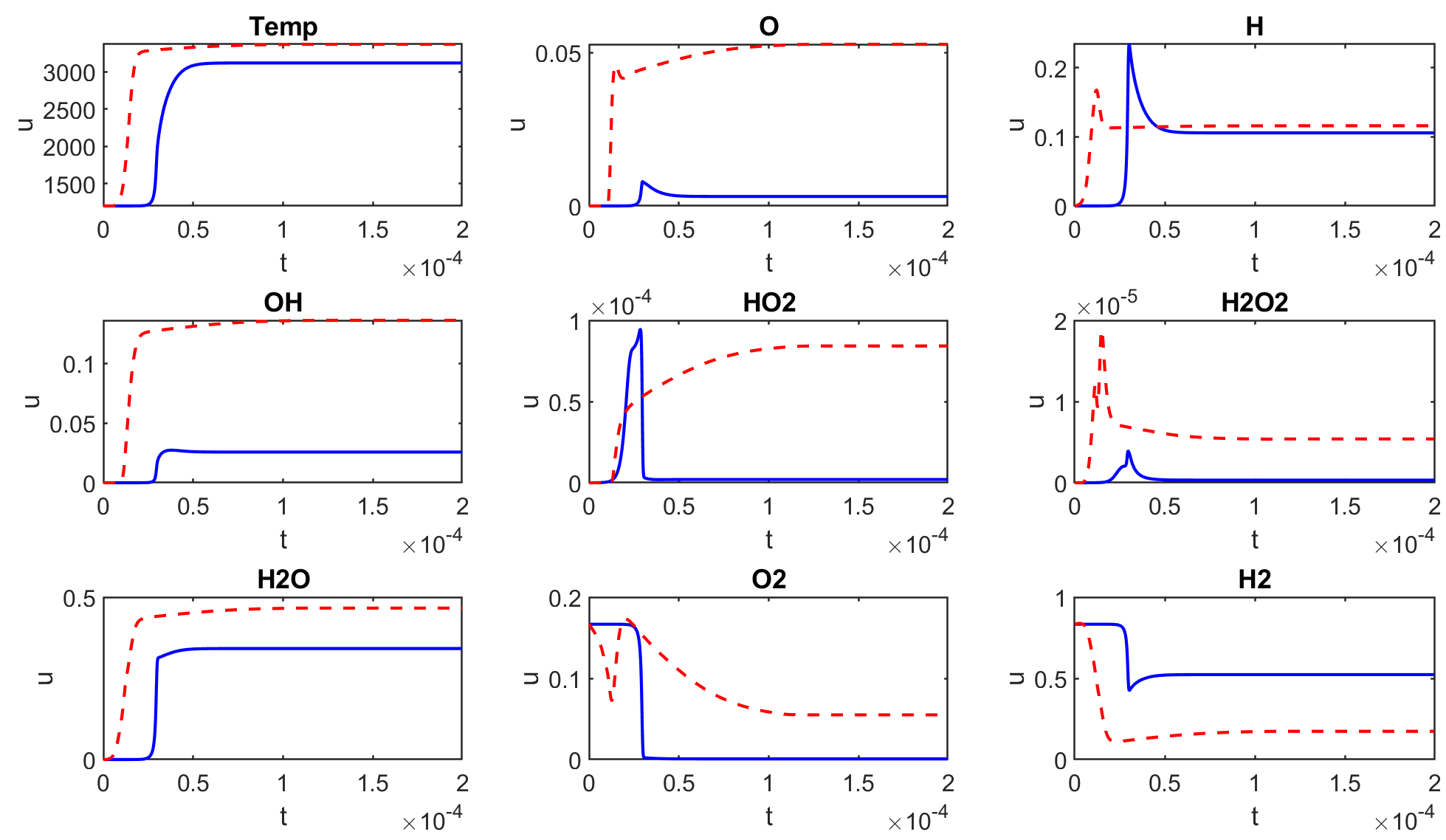}
	
	\caption{Results from using parallel ResNets to predict a single step (left 9 plots) and marching in time from an initial condition (9 right plots) using DNNs with a depth of 7 and a width of 50.}
	\label{fig:earlyNoDer}
\end{figure}

The next result uses the \emph{second approach} described in 
section \ref{sec:Approach}, in which the ResNets are trained to learn 
the next time step rather than the right-hand-side of the ODE.  For the 
results shown in Figure \ref{fig:earlyNoDer} parallel ResNets of depth 7 and 
width 50 were used.  These ResNets were trained using validation data with 
a patience of 400.  For the nine subplots on the left side of 
Figure \ref{fig:earlyNoDer}, the ResNets are being used to predict only a 
single timestep. 
%rather than march in time from an initial condition.  
In other words, the red dashed curves are produced by 
plotting $\widehat{u}^{n+1} = \widehat{S}(u^n)$ for all $n$ where $u^n$ 
comes from known data. The right nine subplots show 
$\widehat{u}^{n+1} = \widehat{S}(\widehat{u}^n)$ where only $u^0$ is 
from known data.
%whereas the main goal of this work, and what is plotted in the right nine subplots, is to generate $\widehat{u}^{n+1} = \widehat{S}(\widehat{u}^n)$ where only $u^0$ is from known data. 
% Even though this is not our primary goal, we 
The plots on the left are included as evidence that the networks are close 
to learning the correct curves.  The accumulation and propagation of errors 
when the quantities are marched in time can significantly affect the accuracy 
of the results as shown in the right plots.  The results shown in 
Figure \ref{fig:earlyNoDer} come from data on which the ResNets were not 
trained.

\begin{figure}[htb]
	\centering
	\includegraphics[width=0.48 \textwidth]{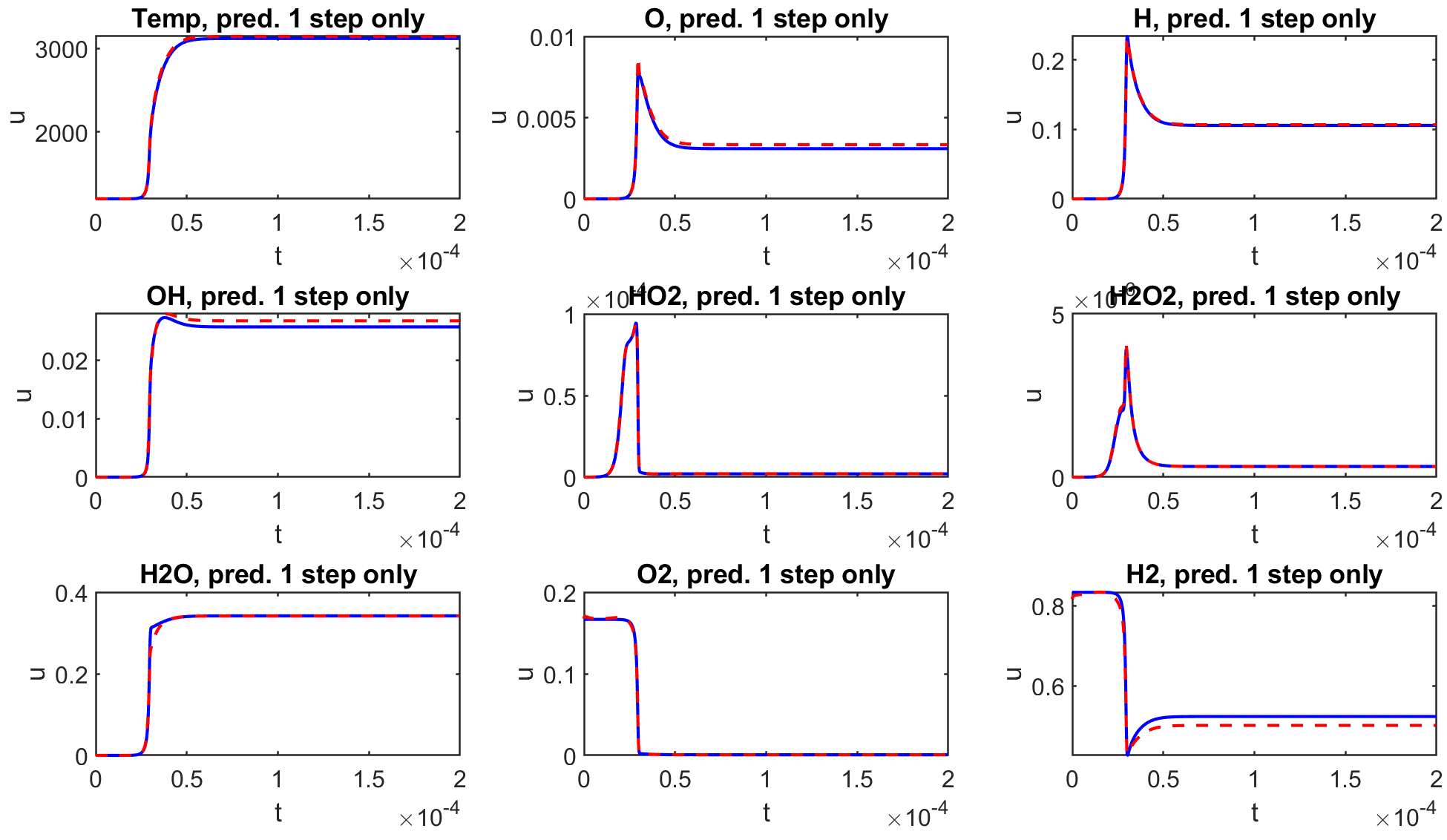} \quad   
 	\includegraphics[width=0.48\textwidth]{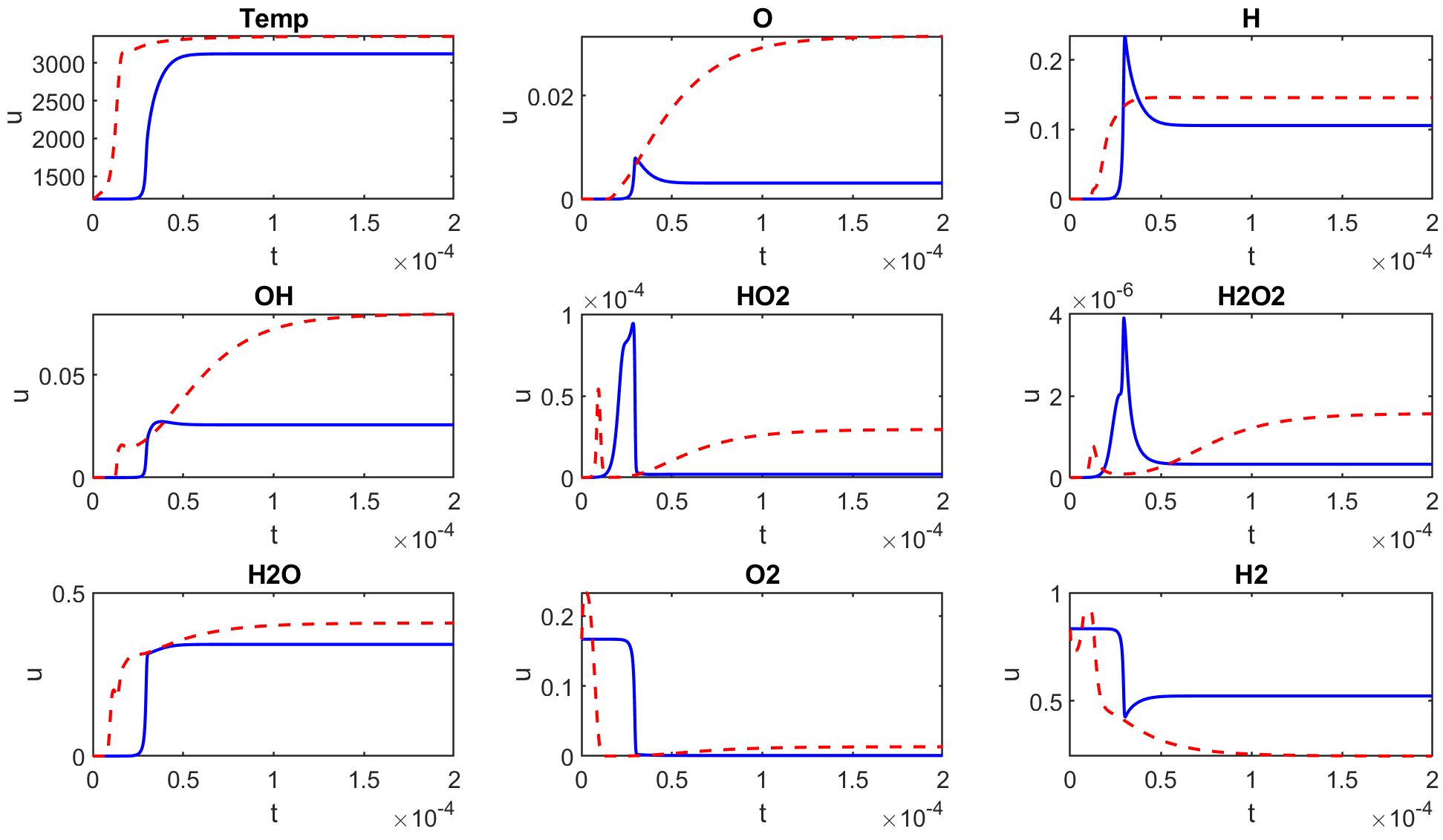}	
	\caption{Predicting a single timestep (left 9 plots) and marching in time (right 9 plots) from ResNets trained on log-scaled data.  The ResNets used here have a depth of 9 and a width of 20, and were trained with validation data with patience of 400 iterations.}
	\label{fig:logResult}
\end{figure}

In Figure \ref{fig:logResult}, results are displayed in which the training 
data was log-scaled (i.e. where $\log x_i$ is used rather than $x_i$ 
in \eqref{eq:scale}). The ResNets used for these results all have depth 9 
and width 20 and were trained using validation data with a patience of 400 
iterations. The plots in Figure \ref{fig:logResult} were created using the 
same data set that was used to create the curves in 
Figure \ref{fig:earlyNoDer}.  From these results, and many others not shown 
here, it is unclear if log-scaling provides any improvement in accuracy. 

\subsection{Grouping training data based on equivalence ratio.} 

\begin{figure}[htb]
	\centering
	\includegraphics[width=0.48 \textwidth]{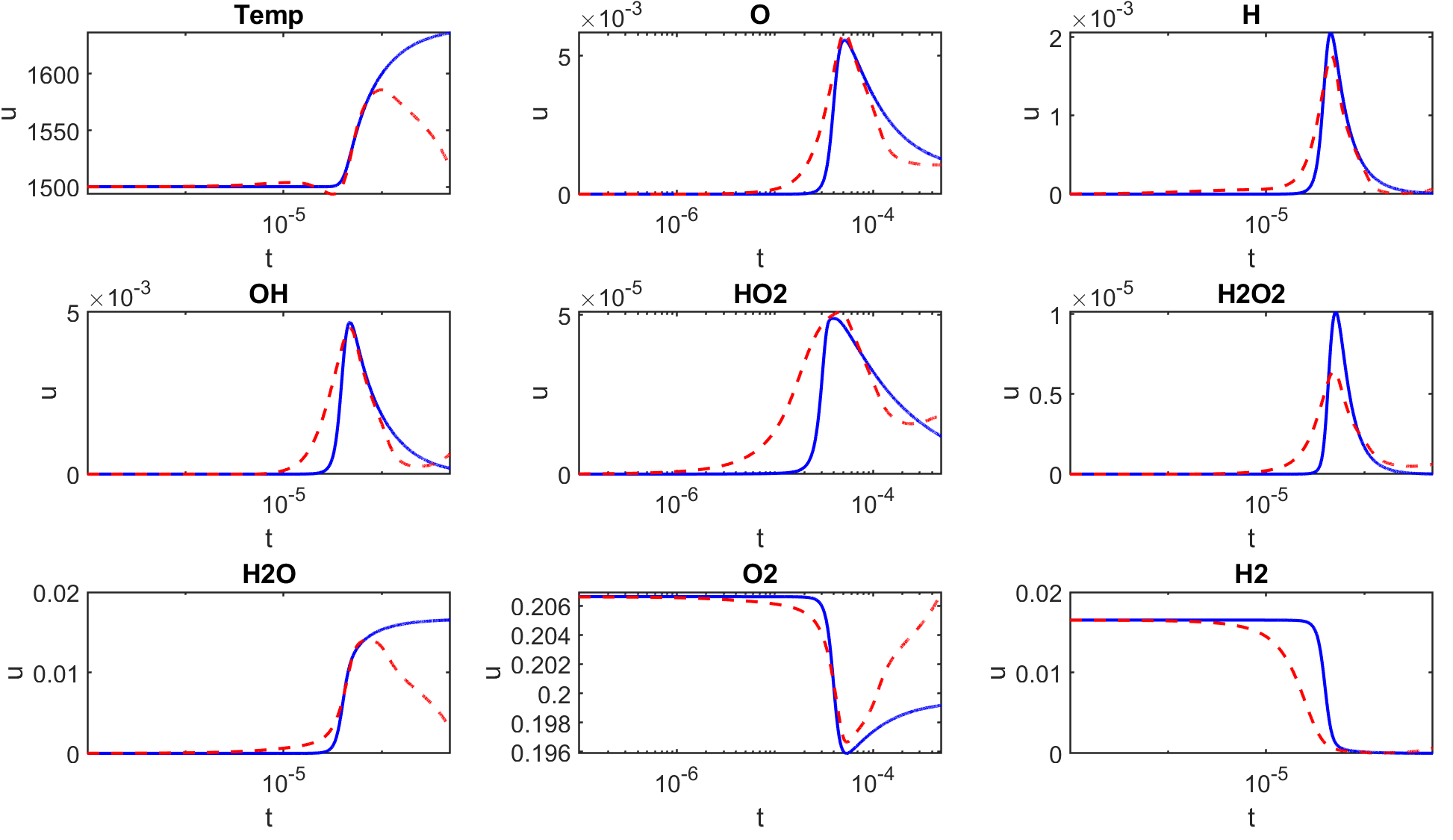} \quad   
 	\includegraphics[width=0.48\textwidth]{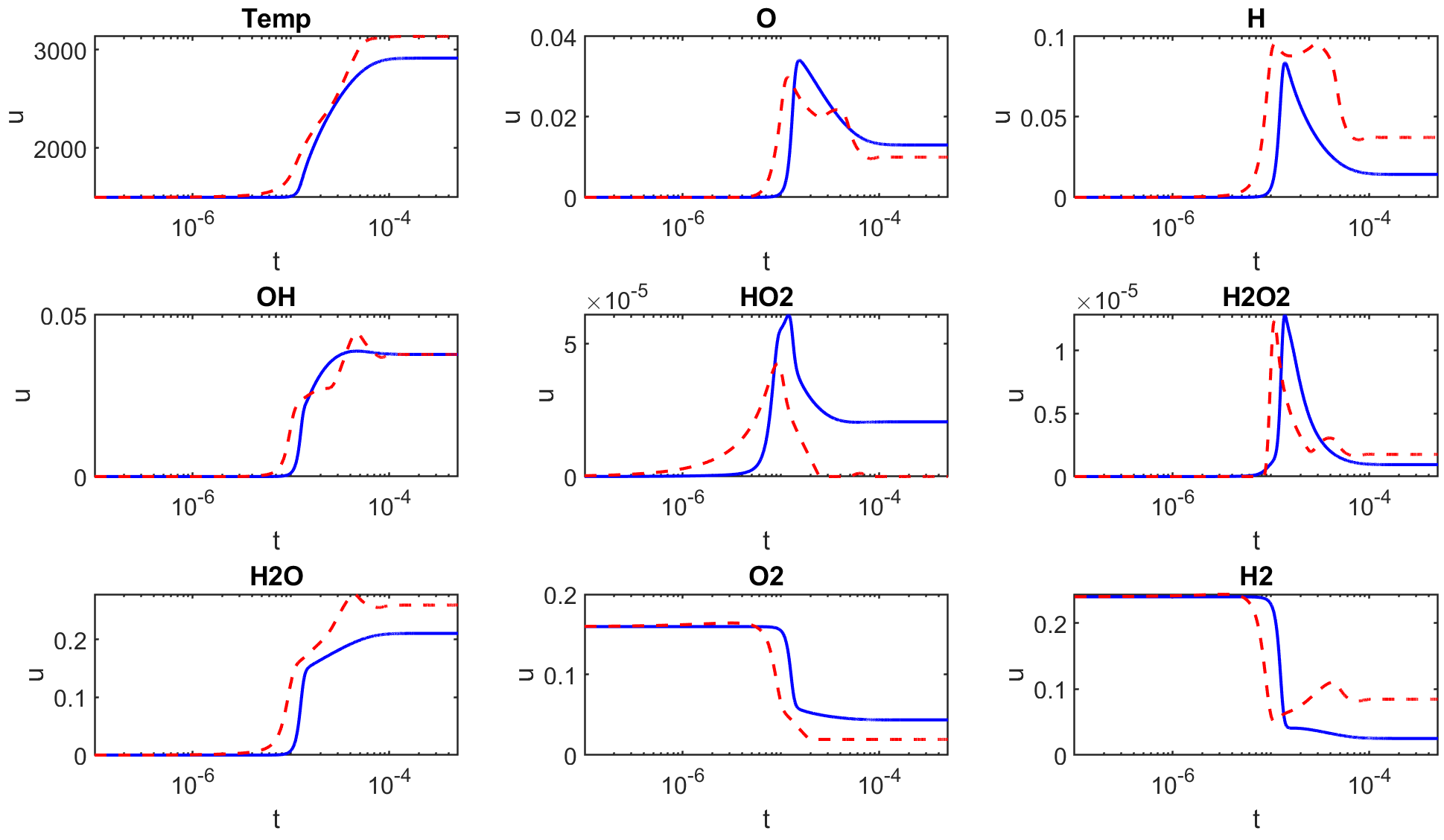}	\\
 	\includegraphics[width=0.48\textwidth]{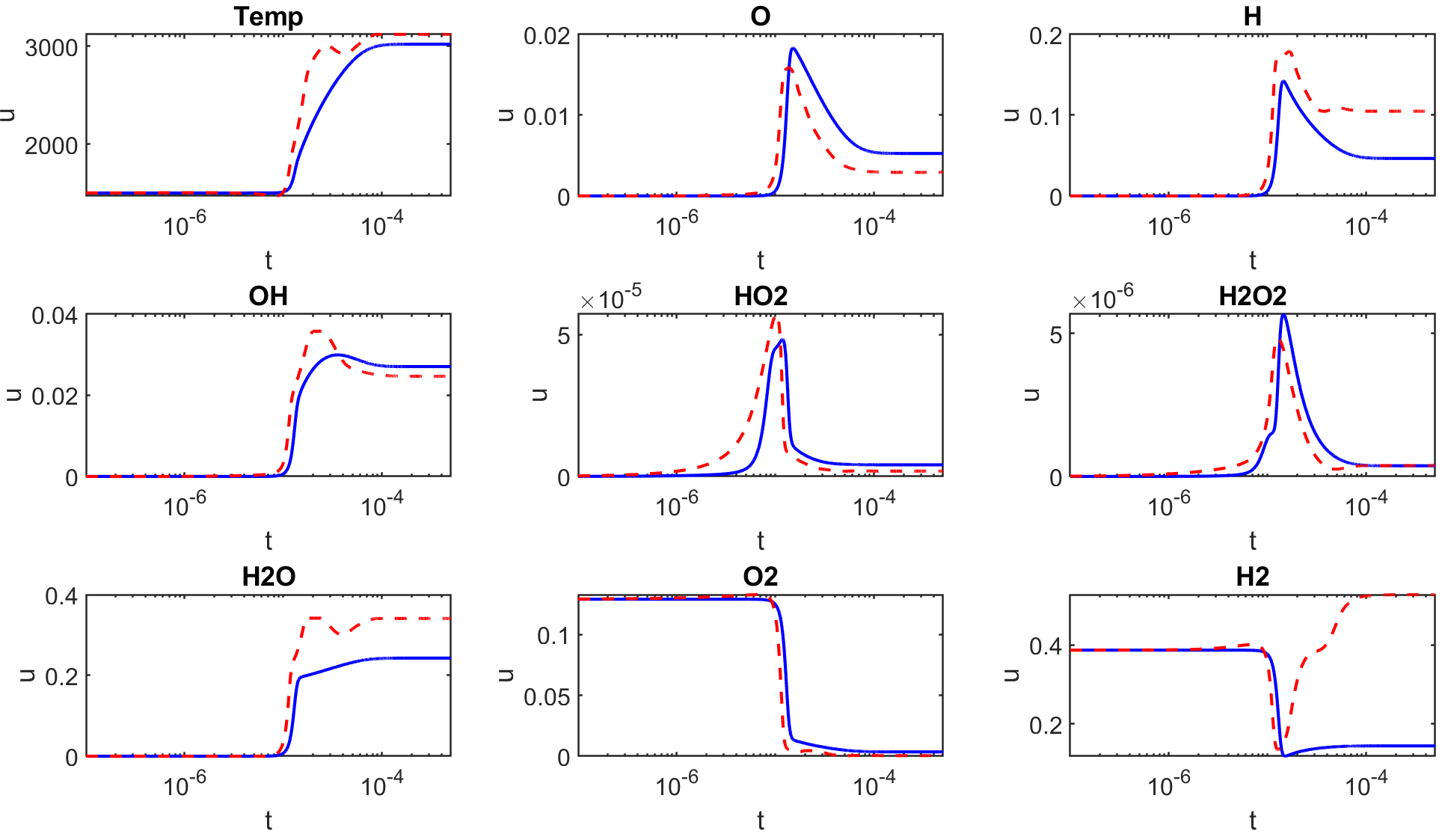}
	\caption{Results from marching initial conditions forward in time using ResNets trained on fuel lean data (top left), fuel balanced (top right), and fuel rich (bottom) data sets.  The ResNets used have a depth of 6 and a width of 30, and were trained with validation data with patience of 400 iterations.}
	\label{fig:fuelGrp1}
\end{figure}

For the next results the training data is split into three groups: fuel 
lean (equivalence ratio $\leq 0.1$), fuel balanced (equivalence ratio 
between 0.1 and 2), and fuel rich (equivalence ratio $> 2$). 
See Appendix \ref{sec:app} for more details on these data sets.  
The results of an experiment using the {second approach}, where a 
different group of parallel ResNets is trained for each of the fuel-based 
data sets, are presented in Figure \ref{fig:fuelGrp1}.  In other words, 
there are 9 plots corresponding to results from ResNets trained only on 
fuel lean data sets, 9 plots for ResNets trained only on fuel balanced 
data sets, and 9 plots for ResNets trained only on fuel rich data sets. 
All ResNets have a depth of 6 and width of 30 and were trained using 
validation data with a patience of 400 iterations.  Furthermore, these 
plots are presented with a logarithmically scaled $x$-axis, as for some 
of the data the reactions happen very quickly.   All of the plots in 
Figure \ref{fig:fuelGrp1} show results of marching in time given only an 
initial condition from known data that was not seen by the networks during 
training.  These results show that the ResNets trained on fuel-based grouping 
is more successful than the previous attempts.   

\subsection{Training data sets with varying initial temperatures, but fixed equivalence ratio.}

\begin{figure}[htb]
	\centering
	\includegraphics[width=0.48 \textwidth]{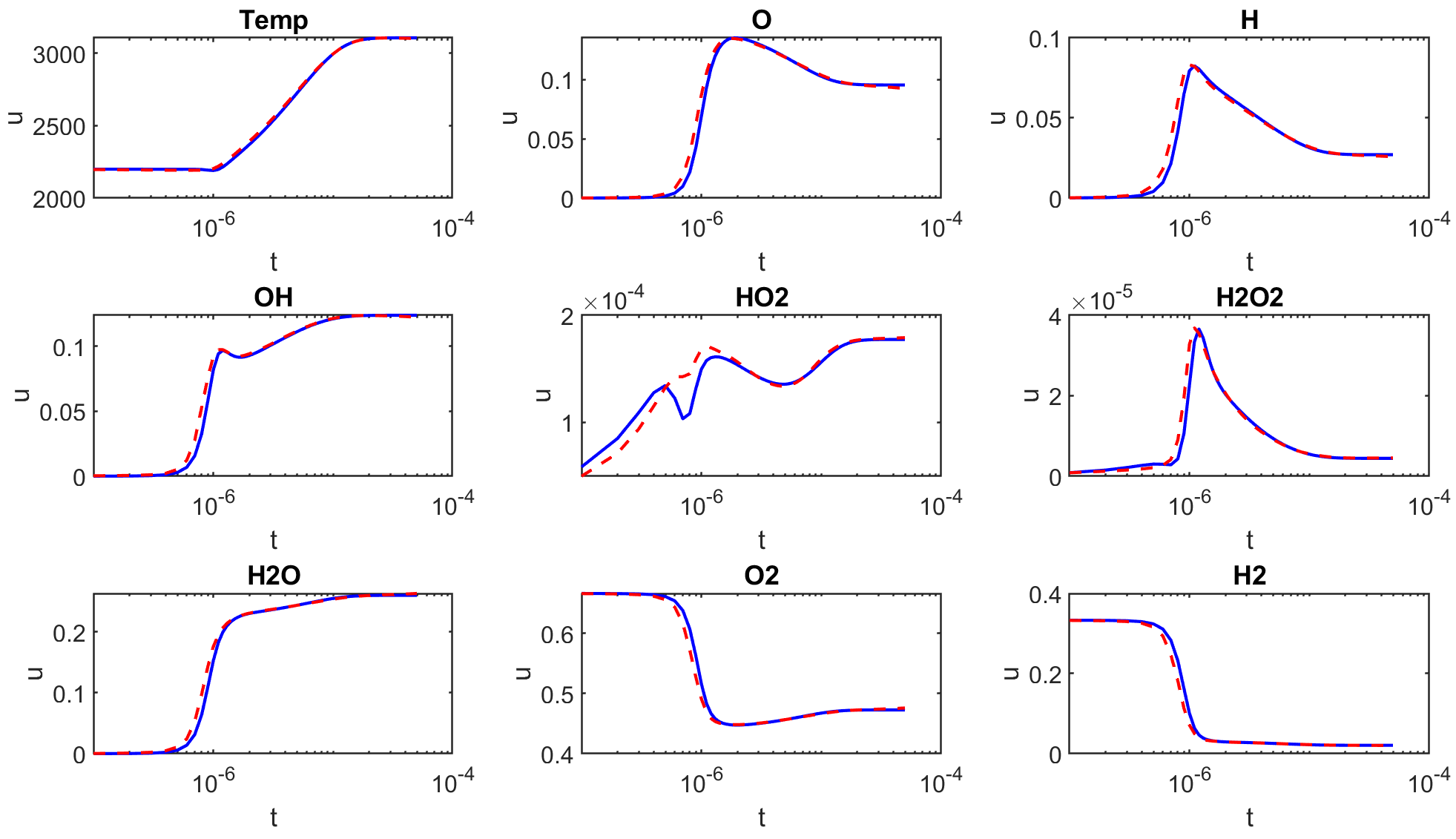} \quad   
 	\includegraphics[width=0.48\textwidth]{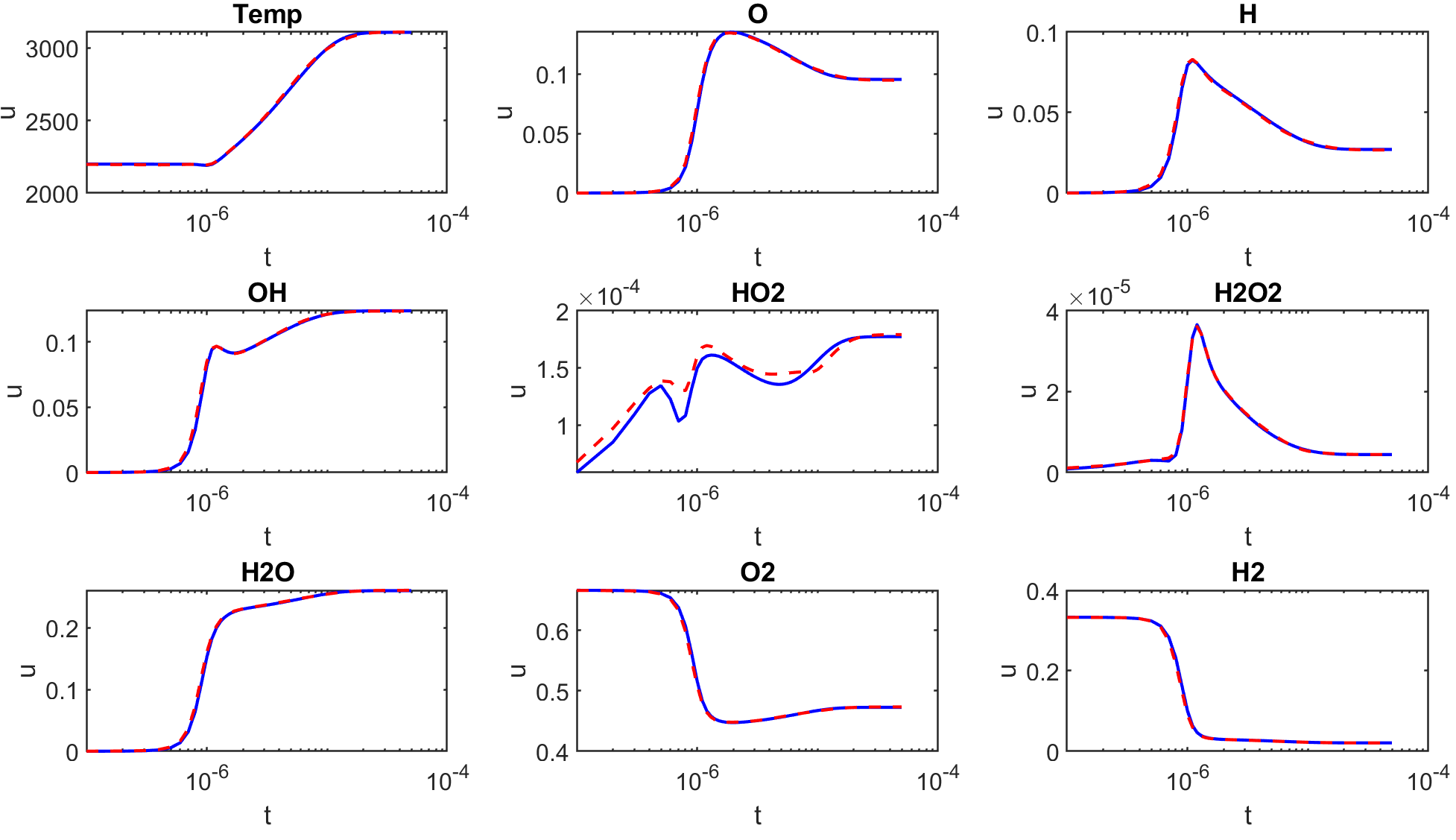}	
	\caption{Results from ResNets trained on data that varies the initial temperature.  The plots on the left were produced from ResNets with a depth of 6 and a width of 30, and were trained with validation data with patience of 400 iterations. The plots on the right came from ResNets with varying depths all with a width of 30, and were trained with validation data with patience of 400 iterations. All plots show the results of marching in time (known data in blue, ResNet results in red).}
	\label{fig:temp}
\end{figure}

For all of the results shown in Figures \ref{fig:earlyParSub} 
through \ref{fig:fuelGrp1} the ResNets were trained on data sets where 
the equivalence ratio was varied and there were at most two different initial 
temperatures (see Appendix \ref{sec:app} for more details).  For this final 
result, parallel ResNets were trained on data sets that had initial conditions 
with the same equivalence ratio, but different initial temperatures.  The 
results of an experiment using DNNs trained on this data can be seen in 
Figure \ref{fig:temp}.  The 9 plots on the left show results from ResNets 
with a depth of 6 and width of 30.  On the right, the results came from 
ResNets with different depths (3 to 8) all with width 30. All of the 
ResNets were trained with validation data with patience of 400 iterations.

\section{Conclusions and Future Directions} \label{sec:Con}

A number of ResNet based approaches to approximate the stiff ODEs arising
in chemical kinetics were developed.
The experiments presented indicate that when the equivalence ratio is 
fixed and initial temperature is varied (this case has been considered
in \cite{ShJoKeMo2020,OwPa2021}), the proposed DNNs can (almost) perfectly
generalize to unseen data. This capacity for generalization deteriorates, 
however, when the initial temperature is kept fixed and the equivalence 
ratio is varied. In order to overcome this issue, the training data was
separated into fuel lean, fuel balanced, and fuel rich based on the 
equivalence ratio. This approach has led to encouraging results.

There are several questions that are currently being investigated:
\begin{itemize}
\item All DNNs are dependent on the quality of training data. The data 
used for experiments reported here were generated using CHEMKIN. In view
of the high dimensionality of the data, and the nature of chemical reactions
(large periods of inactivity followed by a sudden reaction followed by
a convergence to a steady state), quality criteria need to be developed
to make sure redundant data is avoided.
\item It will be interesting to see if further dividing the training 
data based on equivalence ratio is helpful.  		
\item The usefulness of training with noisy data is also being explored. 
For some classification problems, this approach is known to increase
the robustness of ResNets.
\item It is of interest to see how the proposed approach generalizes to 
more reactions and species.
\end{itemize}

\section{Acknowledgements} \label{sec:Ack}
This work was supported by the Defense Threat
Reduction Agency (DTRA) under contract HDTRA1-15-1-0068.
Jacqueline Bell served as the technical monitor. \\
It is a pleasure to acknowledge a number of fruitful discussions 
with Drs. Adam Moses, Alisha J. Sharma and Ryan F. Johnson 
from the Naval Research Laboratory that lead to several improvements
in the techniques discussed in this paper.

\bibliographystyle{abbrv}
\bibliography{references}

\appendix
\section{Description of the Training Data Sets} \label{sec:app}

In this appendix the data sets that were used to train the DNNs and create 
the plots presented above are described in more detail.  Specifically, 
information about the number of points, the equivalence ratio, and initial 
temperature of the individual data sets is provided.  For the results in 
Figure \ref{f:chem1}, a single ResNet was trained with a set containing 
data from a single initial condition.  This set has 4,999 points, and the 
initial condition has equivalence ratio 0.1 and initial temperature 1,200$^o$K.  
Once the network was trained on this data, it was tested with the same 
initial point from the set on which it was trained.  

The parallel ResNets used to produce the plots in 
Figures \ref{fig:earlyParSub} through \ref{fig:logResult} were trained 
on a data set with subsets that correspond to 9 different initial conditions, 
all with initial temperature 1,200$^o$K.  The data is described further in 
Table \ref{tab:1}, where the subsets are numbered for convenient reference. 
This set also contains the data used to create Figure \ref{f:chem1} (set 3).  
After being trained on this entire data set, the ResNets that were used 
for the results in Figure \ref{fig:earlyParSub} were then tested on set 5, 
and therefore these results do not reflect the ResNets generalizing to unseen 
data. 

The subsets of data described in Table \ref{tab:1} are of different sizes.  
For the results in Figure \ref{fig:earlyValDat} the differences in sizes was 
compensated by training with copies of the smaller sets.  Specifically, 
the ResNets were trained with set 1, two copies of sets 3 and 9, and eight 
copies of sets 5 and 7. The ResNets were then tested by using the initial 
condition from set 6, and these are the results that are shown in 
Figure \ref{fig:earlyValDat}.    To train the ResNets that produced the 
results in Figures \ref{fig:earlyNoDer} and \ref{fig:logResult} a single 
copy of sets 1, 3, 5, 7, and 9 from Table \ref{tab:1} was used.  
The trained networks were then tested on set 8. The difference between these 
two experiments are the architecture of the ResNets and the way that the 
training data was scaled prior to training.  Also, as opposed to the 
experiments represented by Figures \ref{fig:earlyParSub} 
and \ref{fig:earlyValDat}, the experiments that resulted in 
Figures \ref{fig:earlyNoDer} and \ref{fig:logResult} use the 
\emph{second approach}, where the ResNets are being trained to learn the 
next timestep of the data.  To create the training data in this situation, 
the same data is used as both components (input and output).  This results 
in size of each subset being reduced by one, because 
there is no next timestep to learn at the final time. 

\begin{table}[h!]
\begin{tabular}{|l|c|c|c|c|c|c|c|c|c|}
\hline
Set & 1 & 2 & 3 & 4 & 5 & 6 & 7 & 8 & 9\\
\hline
Number of points & 8,000 & 6,000 & 4,999 & 1,999 & 999 & 999 & 999 & 1,999 & 3,999 \\
\hline
Equivalence eatio & 0.01 & 0.05 & 0.1 & 0.25 & 0.5 & 1 & 2 & 5 & 10 \\
\hline 
Initial temperature & 1,200 & 1,200 & 1,200 & 1,200 & 1,200 & 1,200 & 1,200 & 1,200 & 1,200 \\
\hline 
\end{tabular}
\caption{Description of the training data used to produce Figures \ref{fig:earlyParSub} through Figure \ref{fig:logResult}.} \label{tab:1}
\end{table}

In Table \ref{tab:2} the data sets that have been separated into fuel lean, 
balanced, and rich sets are described.  The first thing to notice about 
this data is that it contains the data sets from Table \ref{tab:1}, where 
here the number of points is one less for the same reason described above.  
This data was used to create the results shown in Figure \ref{fig:fuelGrp1}.  
The ResNets trained on fuel lean sets (top left set of 9 plots) were 
trained with subsets 1, 3, 4, 6, and 9, and then tested with the initial 
condition from set 8.  The ResNets trained on the fuel balanced sets were 
trained on subsets 1, 3, 5, 7, 9, and 12, and subsequently tested with the 
initial condition from set 11.  Finally the networks trained on fuel rich 
sets were trained with subsets 1, 2, 3, 5, and 7, and then tested with 
the initial condition from set 4.

\begin{table}[h!]
\begin{tabularx}{\textwidth}{|l|>{\centering\arraybackslash}X|>{\centering\arraybackslash}X|>{\centering\arraybackslash}X|>{\centering\arraybackslash}X|>{\centering\arraybackslash}X|>{\centering\arraybackslash}X|>{\centering\arraybackslash}X|>{\centering\arraybackslash}X|>{\centering\arraybackslash}X|}
\hline
\multicolumn{10}{|c|}{Fuel Lean Sets}\\
\hline
Set & 1 & 2 & 3 & 4 & 5 & 6 & 7 & 8 & 9\\
\hline
Number of points & 7,999 & 5,999 & 4,998 & 4,999 & 4,999 & 4,999 & 4,999 & 4,999 & 4,999 \\
\hline
Equivalence ratio & 0.01 & 0.05 & 0.1 & 0.01 & 0.02 & 0.04 & 0.06 & 0.08 & 0.1 \\
\hline 
Initial temperature & 1,200 & 1,200 & 1,200 & 1,500 & 1,500 & 1,500 & 1,500 & 1,500 & 1,500 \\
\hline 
\end{tabularx}
\begin{tabularx}{\textwidth}{|l|>{\centering\arraybackslash}X|>{\centering\arraybackslash}X|>{\centering\arraybackslash}X|>{\centering\arraybackslash}X|>{\centering\arraybackslash}X|>{\centering\arraybackslash}X|>{\centering\arraybackslash}X|>{\centering\arraybackslash}X|>{\centering\arraybackslash}X|>{\centering\arraybackslash}X|>{\centering\arraybackslash}X|>{\centering\arraybackslash}X|}
\hline
\multicolumn{13}{|c|}{Fuel Balanced Sets}\\
\hline
Set & 1 & 2 & 3 & 4 & 5 & 6 & 7 & 8 & 9 & 10 & 11 & 12\\
\hline 
Number of points & 1,998 & 998 & 998 & 998 & 4,999 & 4,999 & 4,999 & 4,999 & 4,999 & 4,999 & 4,999 & 4,999\\
\hline 
Equivalence ratio & 0.25 & 0.5 & 1 & 2 & 0.2 & 0.4 & 0.5 & 0.75 & 0.9 & 1 & 1.5 & 2\\
\hline 
Initial temperature & 1,200 & 1,200 & 1,200 & 1,200 & 1,500 & 1,500 & 1,500 & 1,500 & 1,500 & 1,500 & 1,500 & 1,500 \\
\hline 
\end{tabularx}
\begin{tabularx}{\textwidth}{|l|>{\centering\arraybackslash}X|>{\centering\arraybackslash}X|>{\centering\arraybackslash}X|>{\centering\arraybackslash}X|>{\centering\arraybackslash}X|>{\centering\arraybackslash}X|>{\centering\arraybackslash}X|}
\hline
\multicolumn{8}{|c|}{Fuel Lean Sets}\\
\hline
Set & 1 & 2 & 3 & 4 & 5 & 6 & 7 \\
\hline
Number of points & 1,998 & 3,998 & 4,999 & 4,999 & 4,999 & 4,999 & 4,999 \\
\hline
Equivalence ratio & 5 & 10 & 2.5 & 3 & 3.5 & 4.5 & 5 \\
\hline 
Initial temperature & 1,200 & 1,200 & 1,500 & 1,500 & 1,500 & 1,500 & 1,500 \\
\hline 
\end{tabularx}
\caption{Description of the training data with fuel-based grouping that was used for the results in Figure \ref{fig:fuelGrp1}.} \label{tab:2}
\end{table}

For the experiments that produced the plots in Figure \ref{fig:temp}, 
the training data consists of 13 subsets.  Each subset has 499 points and 
an equivalence ratio of 1.   Furthermore, each subset has a different 
initial temperature beginning with 1,200$^o$K and increasing by increments 
of 100$^o$K to 2,400$^o$K.  For the experiment shown in Figure \ref{fig:temp}, 
the ResNets were trained with the sets corresponding to initial temperatures 
1,200$^o$K, 1,500$^o$K, 1,800$^o$K, 2,100$^o$K, and 2,400$^o$K. 
The trained ResNets were then tested using the initial condition with 
temperature 2,200$^o$K.

\end{document}